\documentclass[runningheads]{llncs}
\usepackage[T1]{fontenc}
\usepackage{graphicx}
\usepackage[math]{blindtext}
\usepackage{tabularx}
\usepackage{amsmath, amssymb, amsfonts}
\usepackage{bm} 
\usepackage{mathtools} 
\usepackage{algorithm}
\usepackage{algorithmic}
\usepackage{hyperref}

\usepackage{subcaption}

\begin{document}
\title{Unsupervised Symbolic Anomaly Detection}
%
%
%
%
%
\author{
Md Maruf Hossain\inst{1}\orcidID{0000-0002-3256-7701} \thanks{Equal contribution.} \and
Tim Katzke\inst{1,2}\orcidID{0009-0000-0154-7735}\protect\footnotemark[1]  
\and
Simon Klüttermann\inst{1}\orcidID{0000-0001-9698-4339} \protect\footnotemark[1] \and
Emmanuel Müller\inst{1,2}\orcidID{0000-0002-5409-6875}}

\authorrunning{M. Hossain et al.}

%
\institute{TU Dortmund University, Dortmund, Germany\and
Research Center Trustworthy Data Science and Security, UA Ruhr, Germany\\
\email{mdmaruf.hossain@tu-dortmund.de, tim.katzke@tu-dortmund.de, Simon.Kluettermann@cs.tu-dortmund.de}
}

\maketitle              
\begin{abstract}
We propose SYRAN, an unsupervised anomaly detection method based on symbolic regression.
Instead of encoding normal patterns in an opaque, high-dimensional model, our method learns an ensemble of human-readable equations that describe symbolic invariants: functions that are approximately constant on normal data.
Deviations from these invariants yield anomaly scores, so that the detection logic is interpretable by construction, rather than via post-hoc explanation.
Experimental results demonstrate that SYRAN is highly interpretable, providing equations that correspond to known scientific or medical relationships, and maintains strong anomaly detection performance comparable to that of state-of-the-art methods.

\end{abstract}
\section{Introduction}
Anomaly detection (AD) is an important problem in machine learning, with countless applications ranging from fraud and fault detection~\cite{fraudapl,sean,auto_appl_electionfraud} to medicine \cite{medicalad} and scientific research~\cite{applbio,mikuni}. In this work, we focus on the common one-class setting, where a model is trained only on (predominantly) normal samples and later assigns anomaly scores to unseen data.
Numerous methods have been proposed for this setting, spanning classical statistical and distance-based approaches to modern deep neural network methods~\cite{metasurvey,surveyruff}, but all ultimately aim to capture regularities in normal data and flag deviations from these patterns as anomalies.

However, despite this progress, most modern anomaly detection algorithms do not allow accessing these identified patterns. Instead, these are usually hidden away in a non-transparent model with thousands to millions of parameters. 
We believe this to be a significant limitation, as anomaly detection is often applied in high-risk scenarios (e.g., healthcare~\cite{medicalad} or predictive maintenance~\cite{sean}), where ensuring that an anomaly detection method works as intended is critical. Furthermore, the black-box behavior of anomaly detection models has led to fields like fairness being understudied~\cite{lemanFairness}, which is especially critical as anomaly detection is increasingly used, for example, to preselect candidates for hiring~\cite{semi_auto_appl_hiring}.
Finally, it should not be underestimated how such extracted patterns can be used to further understand existing behaviors. Many scientific discoveries have started out either as patterns in data~\cite{fourthparadigmScienceWithData} or as anomalies towards those~\cite{StoryPositron}.

Motivated by these limitations, we propose \emph{SYRAN}, a method that models normality through \emph{symbolic invariants}.
Formally, SYRAN learns a collection of scalar functions that are approximately constant on normal data and uses their deviations as anomaly scores.
Each function is represented as a symbolic expression relating input features, yielding anomaly scores that are interpretable by construction.
In contrast to post-hoc explanation techniques~\cite{deanshap}, SYRAN directly produces closed-form equations that can be inspected, validated, and, if desired, incorporated into downstream analyses or decision rules.

Our approach is inspired by Emmy Noether's perspective of describing physical systems through conservation laws~\cite{emmynoether}, and is conceptually related to recent work on surrogate anomaly detection, most notably DEAN~\cite{DEAN_ECML}, which learns deep neural networks that are approximately constant on normal data.
SYRAN adopts a similar invariance viewpoint but instantiates the surrogate as symbolic expressions obtained via symbolic regression~\cite{neurosymbolic} rather than as neural networks.
This shift in representation introduces specific challenges which we tackle in this paper: a tendency toward trivial constant solutions, a combinatorial search over expression structures, and the need to balance data fit against expression complexity for interpretability.



We evaluate our method in terms of both explainability and anomaly detection performance, demonstrating unparalleled explainability with only a minor performance tradeoff on common anomaly detection datasets. Additionally, since our models are mathematical equations, they can be easily be used in almost every programming language and on almost every device.
To facilitate reproducibility, our code is available at \href{https://github.com/KDD-OpenSource/SYRAN}{github.com/KDD-OpenSource/SYRAN}.

\section{Related Works} 

\subsection{Symbolic Regression}

Symbolic Regression (SR) seeks to infer mathematical expressions from data, discovering both model structure and parameters simultaneously \cite{RW_Radwan}. Its capacity to yield interpretable and human-readable models has driven extensive research over the past decades \cite{Makke_2024}.
Notable systems, such as Eureqa~\cite{RW_Schmidt}, have demonstrated SR’s ability to rediscover physical laws from experimental data. However, most existing methods operate in a supervised manner: given an input, they search for a function to approximate the output. 
Instead, we are focusing on the unsupervised case (and particularly anomaly detection) here. Given an input, we search for relationships that are fulfilled in the data. This is more complicated than the normal supervised case, since we have no direct goal to optimize, there might be multiple equations, and we are susceptible to trivial solutions (consider the relationship $\sin(x)^2+\cos(x)^2=1$).

\subsection{Unsupervised Anomaly Detection}

Anomaly detection is the task of detecting unusual, unexpected, or exceptional samples that deviate from a normal pattern~\cite{surveyzhao,surveyruff}. Thus, the main challenge of anomaly detection is to understand this normal pattern. Depending on the algorithm used, this normal pattern is encoded differently. For example, an autoencoder~\cite{aean} utilizes neural networks, whereas IForest~\cite{ifor} employs tree ensembles. However, one fundamental limitation is that the normal pattern is encoded in a high number of complex-to-understand parameters. This, for example, hides situations where the wrong type of normal pattern is learned~\cite{deanshap} and limits what can be understood from a trained model. And while explainability methods exist~\cite{deanshap}, these only provide limited explanations after the fact. Instead, we suggest SYRAN, an anomaly detection method that is inherently explainable, as it is built from symbolic equations derived from the data.

\subsection{Symbolic Regression for Anomaly Detection}

Applying symbolic regression towards anomaly detection is a growing and promising research direction. For example, \cite{RW_Mine_Orbital_SC} uses symbolic regression to detect anomalous orbital trajectories. Similarly, \cite{RW_Mine_Fraud_detection_deep_SC} uses a symbolic approach to detect fraudulent financial transactions, and \cite{RW_Mine_Fault_SC} finds machine failures using symbolic anomaly detection. However, these approaches are far from being general anomaly detection methods, as each of them relies heavily on the attributes of their specific application area. Furthermore, none of these methods employs unsupervised symbolic regression. Instead, they either require access to rare, costly labeled anomalies or simply predict a key metric. 
To the best of our knowledge, our method, SYRAN, is the first general symbolic anomaly detection method that does not require labeled data and can be applied to any application.




\section{The SYRAN Model}  
\label{sec:syran}

Let the training data be denoted as $
X_{\text{train}} = \{\mathbf{x}^{(n)}\}_{n=1}^{N} \subset \mathbb{R}^d
$, which we assume to consist (almost) exclusively of normal samples.
At test time, the goal is to assign an anomaly score $\text{score}(\mathbf{x}) \in \mathbb{R}$ to any $\mathbf{x} \in \mathbb{R}^d$, such that higher scores indicate a higher degree of abnormality.

Instead of learning a black-box representation of the normal data distribution, \emph{SYRAN} (SYmbolic Regression for unsupervised ANomaly detection) models normality through \emph{symbolic invariants}.
We call a function $f : \mathbb{R}^d \rightarrow \mathbb{R}$ an invariant if it is (approximately) constant on normal data.
Without loss of generality, we fix this constant to $1$ and aim for
\begin{equation}
    f(\mathbf{x}) \approx 1
    \quad \text{for all } \mathbf{x} \in X_{\text{train}}.
    \label{eq:invariant}
\end{equation}
Following the idea of learning conserved quantities as in DEAN~\cite{DEAN_ECML}, we search for such functions but instead instantiate them as human-readable symbolic expressions.


\subsection{Learning a Single Symbolic Invariant}
\label{subsec:single-invariant}

\subsubsection{Invariance Loss on Normal Data}

For a candidate invariant $f : \mathbb{R}^d \to \mathbb{R}$, we measure how well it satisfies~\eqref{eq:invariant} on the training data by the average absolute deviation from $1$:
\begin{equation}
    L_1(f)
    = \frac{1}{N} \sum_{n=1}^{N} \big| f(\mathbf{x}^{(n)}) - 1 \big|.
    \label{eq:l1}
\end{equation}
Minimizing the invariant-learning objective $L_1$ encourages $f$ to be approximately constant on $X_{\text{train}}$.
However, in the symbolic setting in particular, this objective alone easily admits trivial solutions that are useless for anomaly detection.

\subsubsection{Avoiding Trivial Solutions}
\label{subsubsec:noise}

The loss $L_1$ is minimized not only by meaningful invariants of the data manifold but also by functions that are globally constant, or excessively convoluted expressions that evaluate to approximately $1$ everywhere.
Such functions cannot distinguish between normal samples and anomalies.
To discourage these trivial solutions, we introduce a noise contrast term.
We generate an auxiliary set $X_{\text{rnd}} = \{\tilde{\mathbf{x}}^{(\ell)}\}_{\ell=1}^{N_{\text{rnd}}}$ by sampling each feature independently from a uniform distribution over its empirical range:
\[
\begin{aligned}
    \tilde{x}^{(\ell)}_j \sim \mathcal{U}\!\Big(
        \min_{n} x^{(n)}_j,\,
        \max_{n} x^{(n)}_j
    \Big),
    \quad j = 1,\dots,d.
\end{aligned}
\]
On this random background, we define
\begin{equation}
     L_{\text{noise}}(f)
    = \frac{1}{N_{\text{rnd}}}
      \sum_{\ell=1}^{N_{\text{rnd}}}
      \big| f(\tilde{\mathbf{x}}^{(\ell)}) - 1 \big|.
    \label{eq:l2}
\end{equation}
Constant functions that minimize $L_1$ also yield $L_{\text{noise}} \approx 0$.
In contrast, a useful invariant should be specific to the normal data manifold and therefore typically violated on random noise. We encode this preference via a hinge-type penalty that encourages $L_{\text{noise}}$ to be at least a margin $\Delta > 0$ to prevent near-constant solutions while also avoiding that the noise term dominates the optimization.

\subsubsection{Complexity Regularization}
\label{subsubsec:complexity}

To obtain interpretable symbolic invariants, we additionally penalize the complexity of $f$.
Let $c(f)$ denote a non-negative complexity measure of the expression (e.g., proportional to the number and expressivity of nodes in its expression tree).
We define a saturating complexity penalty
\begin{equation}
    L_c(f)
    = \log\bigl( 1 + \log(1 + c(f)) \bigr),
    \label{eq:lc}
\end{equation}
where a hyperparameter $\gamma > 0$ controls the strength of this regularization.
The double logarithm ensures diminishing returns on the complexity penalization, s.t. it does not overwhelm the data fit for moderately complex expressions.

\subsubsection{Combined Objective}

Combining these terms, the loss for a single candidate invariant $f$ is
\begin{equation}
    L(f)
    = L_1(f)
    + \max\bigl(0,\, \Delta - L_{\text{noise}}(f)\bigr)
    + \gamma \, L_c(f).
    \label{eq:total-loss}
\end{equation}
The first term enforces approximate constancy on normal data, the second prevents trivial global constants by contrasting them with random noise, and the third promotes compact and interpretable symbolic expressions.

\subsection{Symbolic Parameterization and Optimization}
\label{subsec:symbolic-optimization}

We parameterize each invariant $f$ as a symbolic expression constructed from a fixed set of operators and functions, such as addition, subtraction, multiplication, division, and common nonlinearities (e.g., $\sin$, $\cos$, $\exp$).
Thus, $f$ can be represented as a rooted expression tree whose leaves are input features and constants, and whose internal nodes are operators.

To minimize $L(f)$ in this non-convex, discrete search space, we employ a simple tree-based evolutionary symbolic regression algorithm~\cite{simonsIsing}.
The optimizer maintains a population of candidate expressions, applies mutation and crossover operators to generate new candidates, and selects those with lower loss, while aiming to maintain high diversity throughout the optimization process.
In principle, any symbolic regression engine capable of evaluating a custom fitness function could be used.

\subsection{Ensembling of Symbolic Invariants}
\label{subsec:ensemble}

A single invariant $f$ typically captures only one simple relation in the data.
Real-world datasets may exhibit multiple approximate invariants, and symbolic regression is inherently stochastic.
To increase robustness and expressiveness, SYRAN therefore learns an ensemble of $M$ invariants $\{f_i\}_{i=1}^{M}$ and aggregates their deviations.

\subsubsection{Feature Bagging}

To induce diversity in the ensemble, we adopt feature bagging~\cite{feature-bagging}.
For each ensemble member $i \in \{1,\dots,M\}$, we sample a subset of $K$ features
\[
    S_i \subseteq \{1,\dots,d\},
    \quad |S_i| = K,
\]
without replacement, and restrict $f_i$ to depend only on the corresponding coordinates $\mathbf{x}_{S_i}$.
We then train $f_i$ by minimizing $L(f_i)$ as in~\eqref{eq:total-loss}, using the same training data but with inputs projected onto $S_i$.
This yields a collection of simple, diverse invariants, each focusing on a different subspace of the feature space.

\subsubsection{Calibration and Aggregation}

Each invariant $f_i$ induces a raw deviation score
\begin{equation}
    d_i(\mathbf{x})
    = \big| f_i(\mathbf{x}_{S_i}) - 1 \big|.
    \label{eq:deviation}
\end{equation}
In practice, the scales of the deviations $d_i(\mathbf{x})$ can vary substantially across ensemble members: some invariants remain very close to $1$ on normal data, while others exhibit larger and possibly heavy-tailed deviations, so a naive average of $d_i$ would be dominated by a few high-variance components.

To obtain a robust aggregate, we normalize each component's deviation by its mean deviation on the training data, denoted $\bar{d}_i$, and pass the result through the logistic sigmoid function $\sigma$, yielding calibrated per-component scores $s_i(\mathbf{x})$.
Finally, SYRAN aggregates these calibrated scores by a simple average,
\begin{equation}
    \text{score}(\mathbf{x})
    = \frac{1}{M} \sum_{i=1}^{M} s_i(\mathbf{x}) \quad
    = \quad \frac{1}{M} \sum_{i=1}^{M} \sigma\!\left(\frac{d_i(\mathbf{x})}{\bar{d}_i}\right),
    \label{eq:score}
\end{equation}
where larger values indicate stronger violations of the learned symbolic invariants and therefore a higher likelihood of being anomalous.

\subsection{Algorithm Overview}
\label{subsec:algorithm}

The complete training and inference pipeline is summarized in Algorithm~\ref{alg:syran}.
During training, SYRAN samples feature subsets, and learns invariants via symbolic regression. The evolutionary process iteratively refines the symbolic expressions using mutation and crossover operations over $G$ generations.
At test time, each sample is evaluated by all invariants and the resulting calibrated per-component scores are averaged. 
Notably, due to the independence of the ensemble member functions, training and inference can be largely parallelized.

\begin{algorithm}[htb]
\caption{SYRAN:}
\label{alg:syran}
\begin{algorithmic}[1]
    \REQUIRE Training data $X_{\text{train}} = \{\mathbf{x}^{(n)}\}_{n=1}^{N}$, Test data $X_{\text{test}} = \{\mathbf{x}^{(n)}\}_{n=1}^{N}$, ensemble size $M$, feature bagging size $K$, noise margin $\Delta$, complexity weight $\gamma$, generations G
    \ENSURE Anomaly scores: $\text{score}(X_\text{test}^{m})$
    
    \FOR{$i = 1$ \TO $M$}
        \STATE Sample feature subset $S_i \subseteq \{1,\dots,d\}$ with $|S_i| = K$
        \STATE Draw random noise  $X_{\text{rnd}}$ over $S_i$
        \STATE Initialize random symbolic expression $f_i^{(0)}$
        \FOR{$g = 1$ to $G$}
            \STATE Evaluate $L(f_i^{(g-1)})$, restricting inputs to $\mathbf{x}_{S_i}$
            \STATE Set $f_i^{(g)}$ based on updating $f_i^{(g-1)}$ with mutation and crossover
        \ENDFOR
        \STATE Compute mean deviation $\bar{d}_i = \frac{1}{N} \sum_{n=1}^{N} \big|f_i(\mathbf{x}^{(n)}_{S_i}) - 1\big|$
        \STATE Compute test deviations $d_i(X_\text{test}^{m}) = \big|f_i(X_\text{test}^{m}) - 1\big|$
        
    \ENDFOR
    \STATE $\text{score}(X_\text{test}^{m})=\frac{1}{M}\sum_{i=1}^M \sigma(d_i(X_\text{test}^{m})/\bar{d}_i)$
\end{algorithmic}
\end{algorithm}

\vspace{-2em}

\section{Rediscovering Kepler’s Third Law}
\label{sec:kepler}

To illustrate SYRAN’s ability to uncover meaningful symbolic invariants, we first consider a small, physics-inspired example.
Kepler’s third law states that for bodies orbiting the same central mass, the square of the orbital period $T$ is proportional to the cube of the semi-major axis $a$ of the orbit.
After rescaling units, this relation can be written in the invariant form
\begin{equation}
    \frac{T^2}{a^3} \approx 1,
    \label{eq:kepler-law}
\end{equation}
i.e., the quantity $T^2 / a^3$ is conserved across different orbits.

We construct a two-dimensional training dataset from published orbital parameters of $13$ bodies orbiting the Sun (planets and dwarf planets), each described by its orbital period $T$ and semi-major axis $a$.
All $13$ samples are treated as normal data and constitute $X_{\text{train}}$ for this experiment.
We apply SYRAN exactly as described in Section~\ref{sec:syran} with input dimension $d = 2$ and feature subset size $K = 2$, so that each invariant may depend on both $T$ and $a$. Each equation is the result of the evolutionary algorithm evaluating $G = 30000$ different options.
A modest complexity weight of $\gamma = 0.1$ and a noise margin of $\Delta = 1$ appear to work well in this example and are used throughout the rest of the paper.

As a measure of success, we inspect the ensemble of learned invariants $\{f_i\}$ and count how many are algebraically equivalent to Kepler’s law, up to simple rearrangements of $T^2$ and $a^3$.
In a typical run, roughly $30\%$ of the learned expressions fall into this class.
Examples of such invariants include
\[
\begin{aligned}
    E_1(T,a) = \frac{(a / T)}{(T / a^2)}, \quad
    E_2(T,a) = \frac{(a/T)^2}{a^{-1}}, \quad
    E_3(T,a) = a \left(\frac{a}{T}\right)^2.
\end{aligned}
\]

This demonstrates that given only a small set of ``normal'' observations and without any prior knowledge of the underlying law, SYRAN can automatically rediscover a compact symbolic invariant that matches a well-known physical relation. While one of the most impactful physicists of history required about 10 years to find this equation, SYRAN can do the same in about one minute of computation.
It thereby provides an interpretable sanity check for our methodology and illustrates it’s potential for data-driven discovery of conserved quantities.

\section{Experiments} 

We evaluate SYRAN on $19$ publicly available datasets from the ADBench benchmark suite~\cite{surveyzhao}, covering diverse real-world domains including healthcare, biochemical, and scientific data.
Restricting attention to those ADBench datasets with at most $30$ features yielded $21$ candidates.
Among these, the fault and wdbc datasets are excluded because their optimization did not reliably complete within a $2$-hour time budget, leaving $19$ datasets: pima, breastw, cardio, stamps, cardiotocography, lymphography, pageblocks, glass, waveform, annthyroid, yeast, pendigits, wilt, hepatitis, wine, thyroid, wbc, vowels, vertebral.
Unless stated otherwise, we use a single set of hyperparameters for all datasets, chosen based on the experiment in Section~\ref{sec:kepler}: complexity weight $\gamma = 0.1$, noise margin $\Delta = 1$, and feature subset size $K = 2$.
For each dataset we train an ensemble of $M = 50$ invariants.
The effect of varying these hyperparameters is analyzed in Section~\ref{sec:hyperparam_effect}.

\subsection{Experimental Results}

\begin{figure}[ht]
\centering
\begin{subfigure}{\textwidth}
\centering
\includegraphics[width=0.99\textwidth]{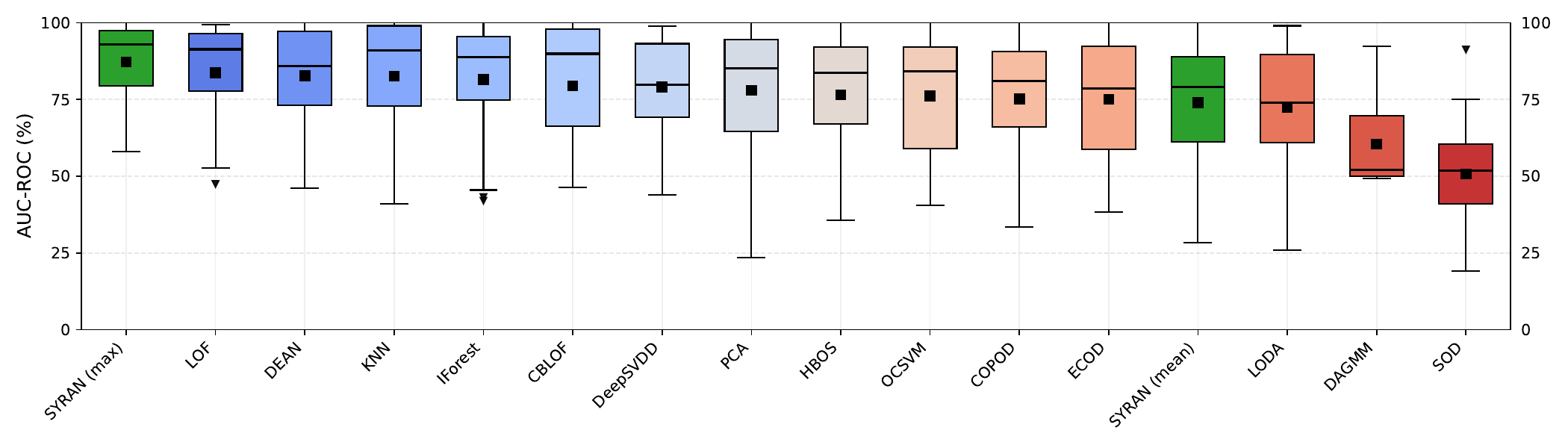}
\caption{AUC-ROC scores for each algorithm across all datasets considered, sorted by mean. For each box, the line indicates the median, and the square the mean performance.}%
\label{fig:auc-boxplots}
\end{subfigure}
\begin{subfigure}{0.49\textwidth}
\centering
\includegraphics[width=\textwidth]{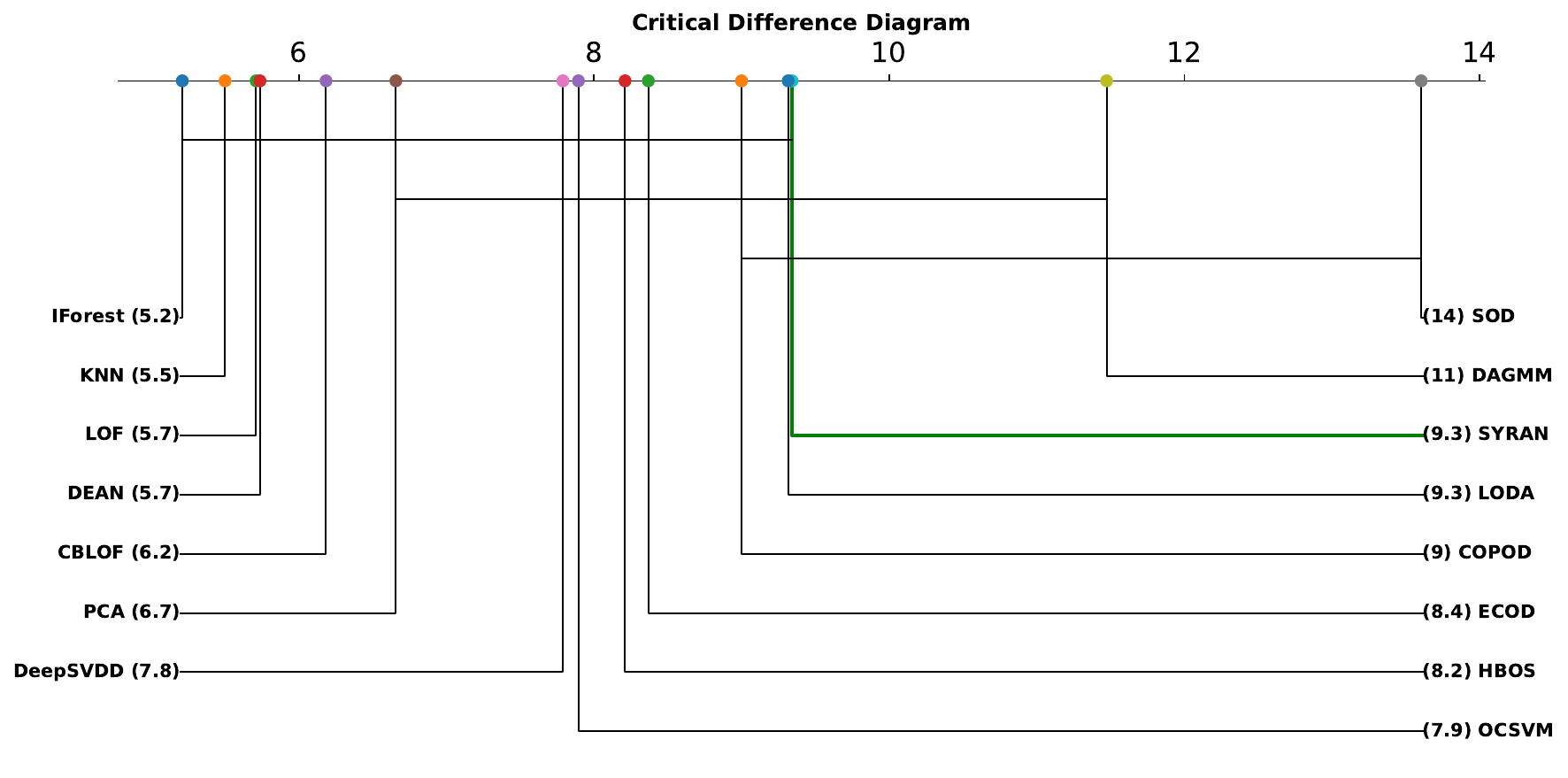}
\caption{CD diagram for mean AUC-ROC.}
\label{fig:cd_mean}
\end{subfigure}
\hfill
\begin{subfigure}{0.50\textwidth}
\centering
\includegraphics[width=\textwidth]{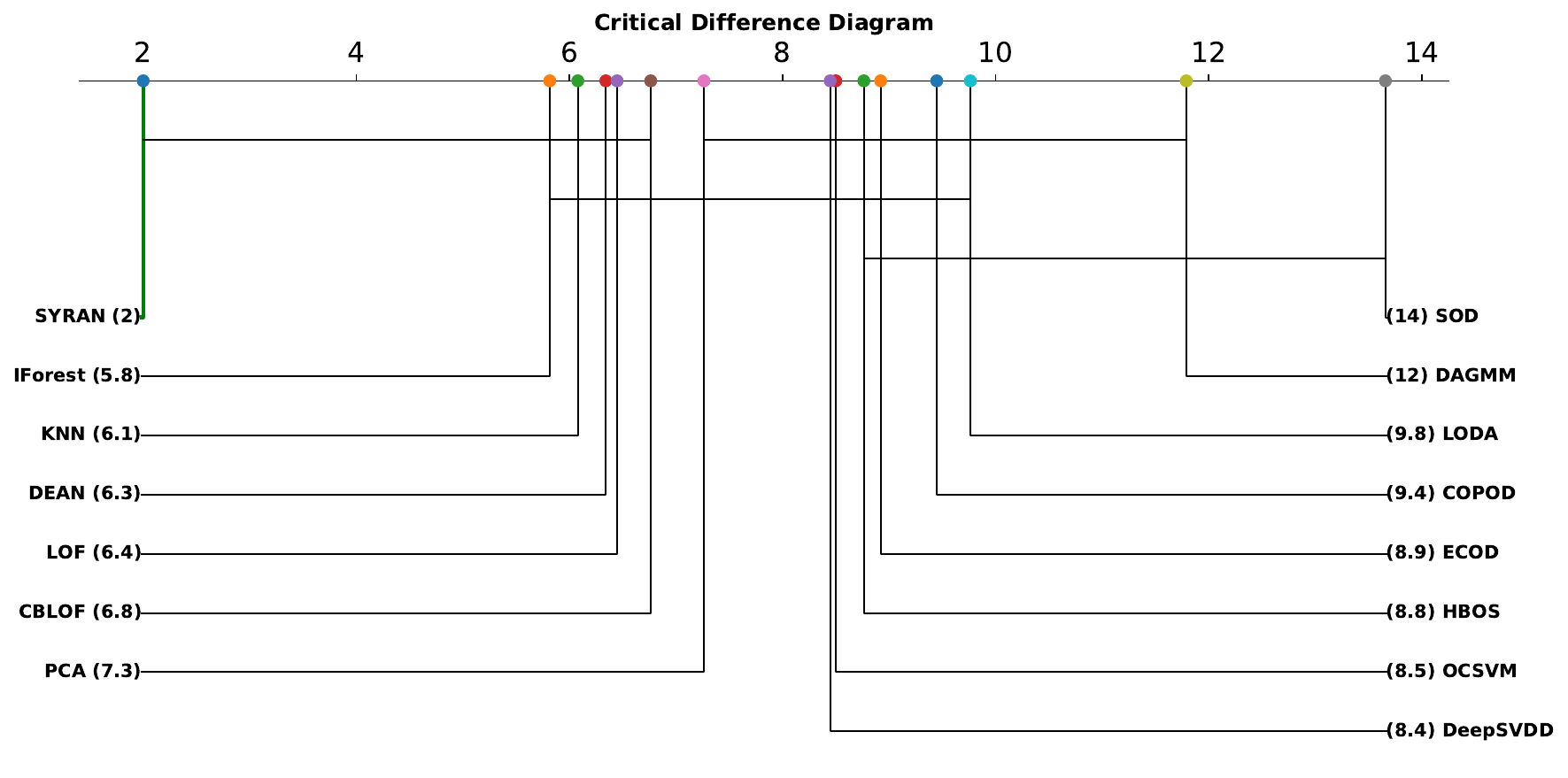}
\caption{CD diagram for max AUC-ROC.}
\label{fig:cd_max}
\end{subfigure}
\caption{
Box plots and critical difference diagrams comparing the AUC-ROC performance of the SYRAN ensemble (a,b) and its best-performing ensemble member (a,c) against baseline anomaly detection methods.}
\label{fig:cd_combined}
\end{figure}


Figure~\ref{fig:cd_combined} presents the performance of our method in relation to competitor results for 14 competitor algorithms, as reported in reference~\cite{DEAN_ECML}, spanning from classical shallow methods to modern deep learning ensembles.

The boxplot in Figure~\ref{fig:auc-boxplots} shows, that the overall ensemble \emph{SYRAN (mean)} achieves competitive performance, while the best individual invariant in each ensemble \emph{SYRAN (max)}, obtained by selecting the ensemble member with highest AUC-ROC per dataset, attains the highest mean performance across datasets.

To assess statistical significance, we follow the standard Friedman--Wilcoxon protocol.
We first apply a Friedman test~\cite{friedman} to detect overall differences between methods and then perform pairwise Wilcoxon tests~\cite{wilcoxon} with Bonferroni--Holm correction~\cite{correction} at $\alpha = 0.05$.
The resulting critical difference (CD) diagrams for the ensemble and the best performing ensemble member are shown in Figures~\ref{fig:cd_mean} and~\ref{fig:cd_max}, respectively. 
For the ensemble, the average rank is slightly worse than that of some baselines (e.g., LOF and DEAN), but the CD diagram indicates that these differences are not statistically significant.
With its best ensemble members alone, however, SYRAN achieves the best average rank and no method significantly outperforms it, and several weaker baselines are significantly worse.

This gap highlights the potential of leveraging interpretability.
Each ensemble consists of both strongly and weakly indicative invariants, but as we will show in Section~\ref{sec:rw}, a domain expert can easily inspect and select the most plausible invariant for a given application, yielding a substantial performance gain over blind averaging.
Among all methods considered, SYRAN is the only one that naturally supports such expert-in-the-loop selection based on closed-form equations.
Moreover, the fact that SYRAN can outperform more complex baselines on several datasets using a single invariant over at most two features suggests that some ADBench tasks may be unsuitable for evaluating methods designed for highly complex data.

\subsection{Interpreting Learned Equations}
\label{sec:rw}

\begin{figure}[t]
  \centering
  \begin{subfigure}[b]{0.32\textwidth}
    \centering
    \includegraphics[width=\linewidth]{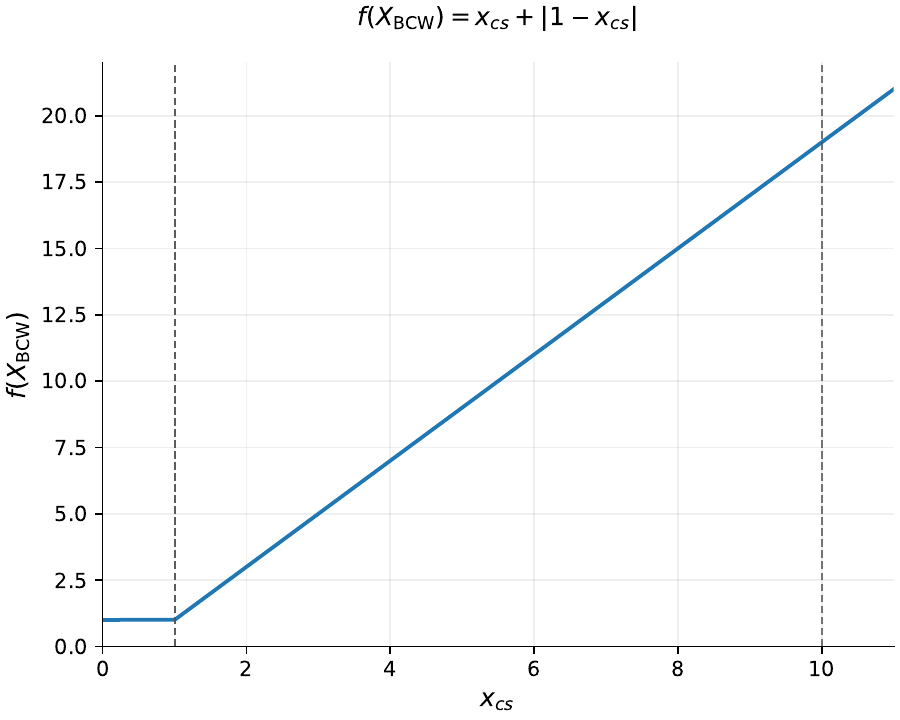}
    \caption{breastw}
    \label{fig:f_BCW}
  \end{subfigure}
  \hfill
    \begin{subfigure}[b]{0.32\textwidth}
    \centering
    \includegraphics[width=\linewidth]{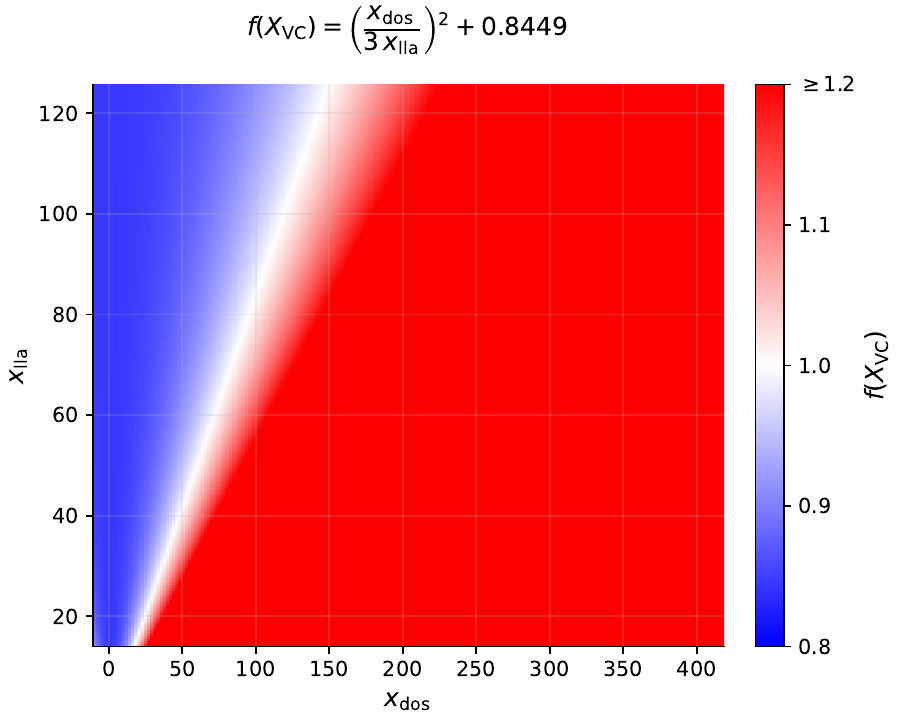}
    \caption{vertebral}
    \label{fig:f_VC}
  \end{subfigure}
  \hfill
  \begin{subfigure}[b]{0.32\textwidth}
    \centering
    \includegraphics[width=\linewidth]{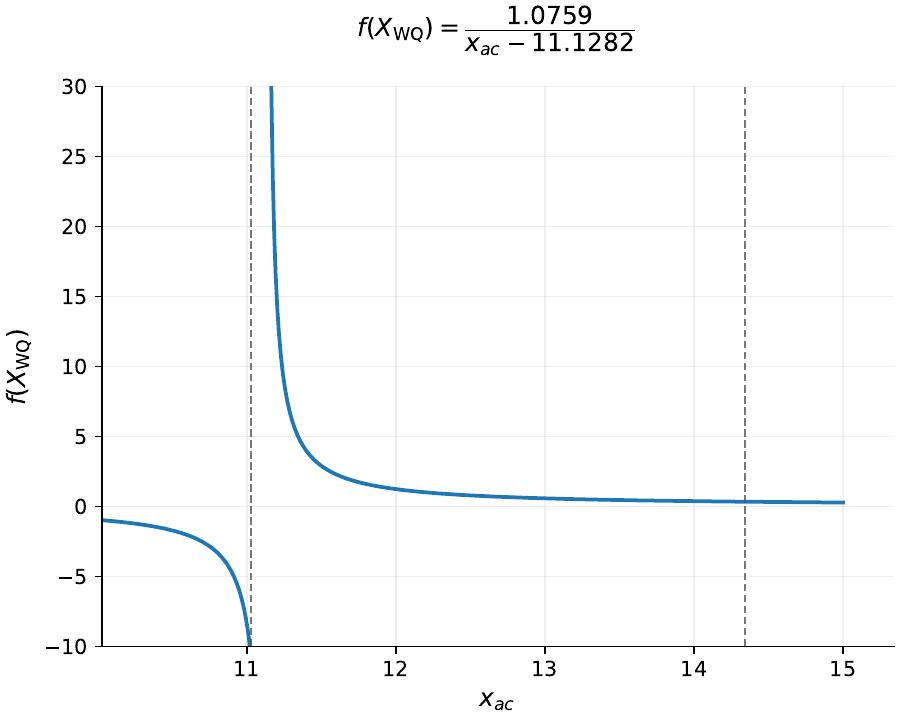}
    \caption{wine}
    \label{fig:f_WQ}
  \end{subfigure}
    \caption{Symbolic invariants learned by SYRAN on three ADBench datasets. Significant deviations of  function values from 1 are indicative of anomalous behavior.
  }
  \label{fig:feature_functions}
\end{figure}

Beyond aggregate performance metrics, SYRAN provides explicit symbolic invariants that can be inspected and interpreted.
In this subsection, we analyze individual invariants learned on three representative datasets from ADBench.
For each dataset, we visualize in Figure~\ref{fig:feature_functions} the behavior of the single ensemble member with the highest AUC-ROC over its feature space.

The \textbf{breastw} dataset~\cite{ds_breastw} concerns breast cancer diagnosis. 
One of its features, $x_{\text{cs}}$, encodes the uniformity of cell size on a discrete scale from $1$ (highly uniform) to $10$ (highly non-uniform), a property plausibly associated with malignancy.
On this dataset, the best invariant depends only on $x_{\text{cs}}$ and has the form
\(
    f_{\text{BCW}}(x_{\text{cs}}) = x_{\text{cs}} + |1 - x_{\text{cs}}|
\),
behaving in line with medical intuition.


The \textbf{vertebral} dataset~\cite{ds_vertebral} contains biomechanical measurements of the lumbar spine for detecting pathological conditions.
Here, the most indicative invariant learned by SYRAN relates the degree of spondylolisthesis $x_{\text{dos}}$ and the lumbar lordosis angle $x_{\text{lla}}$ via
\(
    f_{\text{VC}}(x_{\text{dos}}, x_{\text{lla}})
    = \bigl(x_{\text{dos}}/(3 x_{\text{lla}})\bigr)^2 + 0.8449.
\),
achieving a AUC-ROC of roughly $90\%$ on this dataset.
Although we did not find a direct medical reference for this specific formula, it suggests a plausible joint indicator worthy of further clinical investigation.


Finally, the \textbf{wine} dataset~\cite{ds_wine} describes chemical properties of Italian wines, with one wine type treated as normal and the others as anomalies.
The best-performing learned invariant depends only on the alcohol content $x_{\text{ac}}$ and takes the form
\(
    f_{\text{W}}(x_{\text{ac}}) = 1.0759 / (x_{\text{ac}} - 11.1282).
\)
Within the empirically observed range of $x_{\text{ac}}$, this expression effectively separates normal from anomalous wines, yielding a AUC-ROC of $100\%$.
As we lack detailed documentation on which wine types are designated normal, we cannot confirm whether this matches established oenological knowledge; however, the fact that a single one-dimensional invariant perfectly solves the task raises questions about the usefulness of this benchmark for evaluating anomaly detection methods.


These three cases illustrate how SYRAN's symbolic invariants can recover established domain indicators, propose new candidate relationships between measurements, and expose datasets that are trivially separable by a single feature.

\subsection{Ablation study}
\label{sec:hyperparam_effect}

\begin{figure}[ht]
    \centering
    \includegraphics[width=0.8\textwidth]{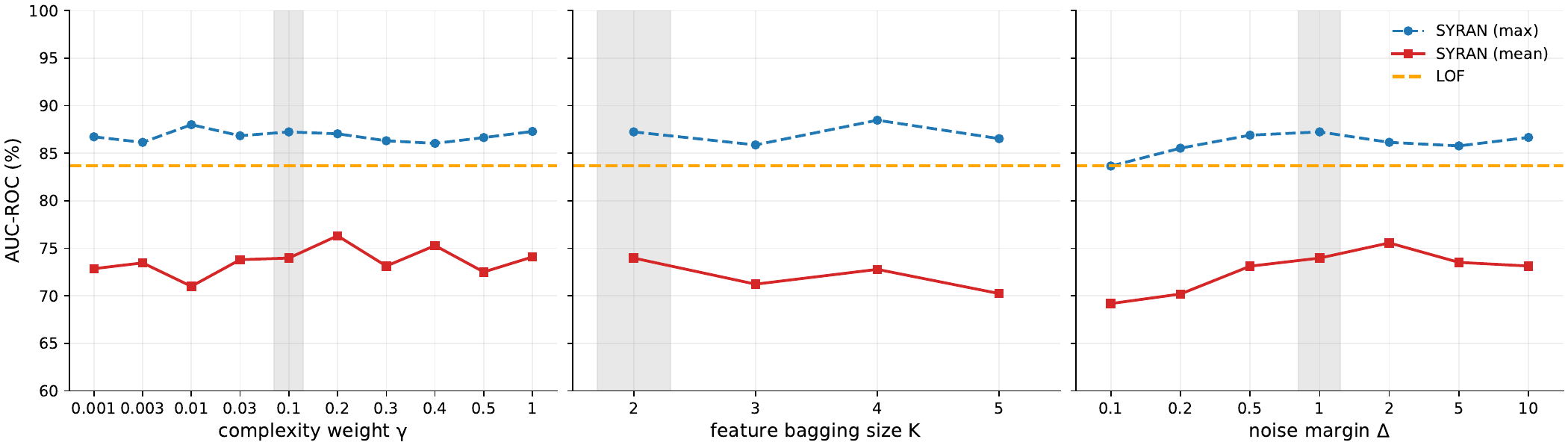}
    \caption{Mean AUC-ROC performance across benchmark datasets for SYRAN and its best ensemble member across different hyperparameterizations compared to LOF as its strongest competitor. The default configuration is highlighted in gray.}%
    \label{fig:hyperparam}
\end{figure}

Our three hyperparameters complexity weight $\gamma$, feature bagging size $K$, and noise margin $\Delta$, affect both interpretability and predictive performance.
While we deliberately chose our benchmark hyperparameters based on the example from Section~\ref{sec:kepler} to ensure fairness, it is likely that they are suboptimal. Thus, we state the performance for alternative paramterizations in Figure~\ref{fig:hyperparam}.
We vary one hyperparameter at a time and leave the remaining two at their default values.
The results are remarkably stable, and as anticipated, there are even noticeable improvements in performance.
Overall, it seems that our choice of default hyperparameters derived from Section~\ref{sec:kepler} is reasonable, but not optimal. For example, by choosing $\gamma=0.03$, $K=2$, $\Delta=2$, the performance of the ensemble changes by $+4.39\%$, and of the best member by $-0.05\%$, leading to substantially better benchmark results. However, we did  not permit such variation of parameters, as this would be an unfair comparison, and they might not generalize.

Next to the anomaly detection performance, another important metric is the interpretability of the equations that we learned. This is mostly affected by the hyperparameter $\gamma$. Thus, we give the best equation on the breastw~\cite{ds_breastw} dataset for various values of $\gamma$ in Table~\ref{tab:complexity_effect}, showing that it is possible to trade between high performance and high expressivity in the equations we learn.

\begin{table}[h]
\caption{Effect of the complexity weight on symbolic expressions for the breastw dataset~\cite{ds_breastw}. Here, $x_i$ represent the $i$-th feature of the dataset ($x_{2}$ corresponds to $x_{cs}$ in Section~\ref{sec:rw}). We state the best ensemble member equation for each $\gamma$.}
\vspace{0.5ex}
\centering
\small
\begin{tabular}{c|c|c}
\hline
\textbf{Complexity weight $\gamma$} & \textbf{Best Equation} & \textbf{AUC-ROC ($\%$)} \\
    \hline
    0.001 & $\left( \frac{1.0500}{x_{3}} \right)^{x_{6} \cdot (-0.0526) \cdot 2.2257}$ & $99.19$ \\
    0.01 & $x_{2}^{\cos\left(\frac{-1.6023}{x_{4}}\right)}$ & $97.60$ \\
    0.10 & $x_{2} + \left| 1 - x_{2} \right|$ & $97.73$ \\
    0.50 & $x_{2}$ & $97.73$ \\
    \hline
\end{tabular}

\label{tab:complexity_effect}
\end{table}



\section{Conclusion} 
SYRAN shows that symbolic regression can serve as a viable and interpretable approach for unsupervised anomaly detection. It learns ensembles of human-readable symbolic invariants that are approximately constant on normal data, optimized in a way that enforces constancy, penalizes trivial constants via random-noise contrast, and controls expression complexity to favor compact equations. 

In experiments, SYRAN rediscovers Kepler’s third law and, on 19 ADBench datasets, achieves AUC-ROC performance competitive with black-box baselines while yielding closed-form anomaly scores that recover known medical indicators, suggest new relationships, and reveal potential flaws in established benchmarks. These properties make SYRAN well suited for expert-guided anomaly detection. Future work includes scaling to higher-dimensional and temporal data and incorporating constraints or priors to improve interpretability and detection accuracy.

\subsubsection{\ackname}
This research was supported by the Research Center Trustworthy Data Science and Security (\url{https://rc-trust.ai}), one of the Research Alliance centers within the University Alliance Ruhr (\url{https://uaruhr.de}).

%
%
%
\bibliographystyle{splncs04}
\bibliography{new_refs}

\end{document}